\documentclass[10pt,twocolumn,letterpaper]{article}

\usepackage{iccv}              

\usepackage[utf8]{inputenc} 
\usepackage[T1]{fontenc}    
\usepackage{hyperref}       
\usepackage{url}            
\usepackage{booktabs}       
\usepackage{amsfonts}       
\usepackage{nicefrac}       
\usepackage{microtype}      
\usepackage{xcolor}         
\usepackage{xspace}

\title{Retrieval Augmented Comic Image Generation}

\author{
  Yunhao Shui,
  Xuekuan Wang,
  Feng Qiu,
  Yuqiu Huang,
  Jinzhu Li,
  Haoyu Zheng,
  \\
  Jinru Han,
  Zhuo Zeng,
  Pengpeng Zhang,
  Jiarui Han,
  Keqiang Sun\\
  Zulution AI
}

\newcommand{\model}{RaCig\@\xspace}
\newcommand{\dataset}{MSDB\@\xspace}
\newcommand{\msi}{HMSI\@\xspace}

\newcommand{\tprompt}{c_{\scriptsize \mbox{text}}}
\newcommand{\cfeature}{f_{\scriptsize \mbox{CLIP}}}
\newcommand{\ipfeature}{f_{\scriptsize \mbox{IP}}}
\newcommand{\closest}{I_{\scriptsize \mbox{nn}}}
\newcommand{\iRef}{I_{\scriptsize \mbox{ref}}}
\newcommand{\db}{D_{\scriptsize \mbox{ms}}}

\newcommand{\closestSkeleton}{S_{\scriptsize \mbox{nn}}}
\newcommand{\closestSegMap}{M_{\scriptsize \mbox{nn}}}
\newcommand{\closestRef}{P_{\scriptsize \mbox{nn}}}
\newcommand{\aaca}{N_{\scriptsize \mbox{asgn}}}
\newcommand{\rci}{N_{\scriptsize \mbox{rci}}}
\newcommand{\ipa}{N_{\scriptsize \mbox{IP}}}

\newcommand{\order}{O}

\newcommand{\ipface}{f_{\scriptsize \mbox{face}}}
\newcommand{\ipbody}{f_{\scriptsize \mbox{body}}}

\newcommand{\iRefface}{I_{\scriptsize \mbox{face}}}
\newcommand{\iRefbody}{I_{\scriptsize \mbox{body}}}

\begin{document}



\twocolumn[
\maketitle
\vspace{-3em}
\begin{center}
  \includegraphics[width=1\linewidth]{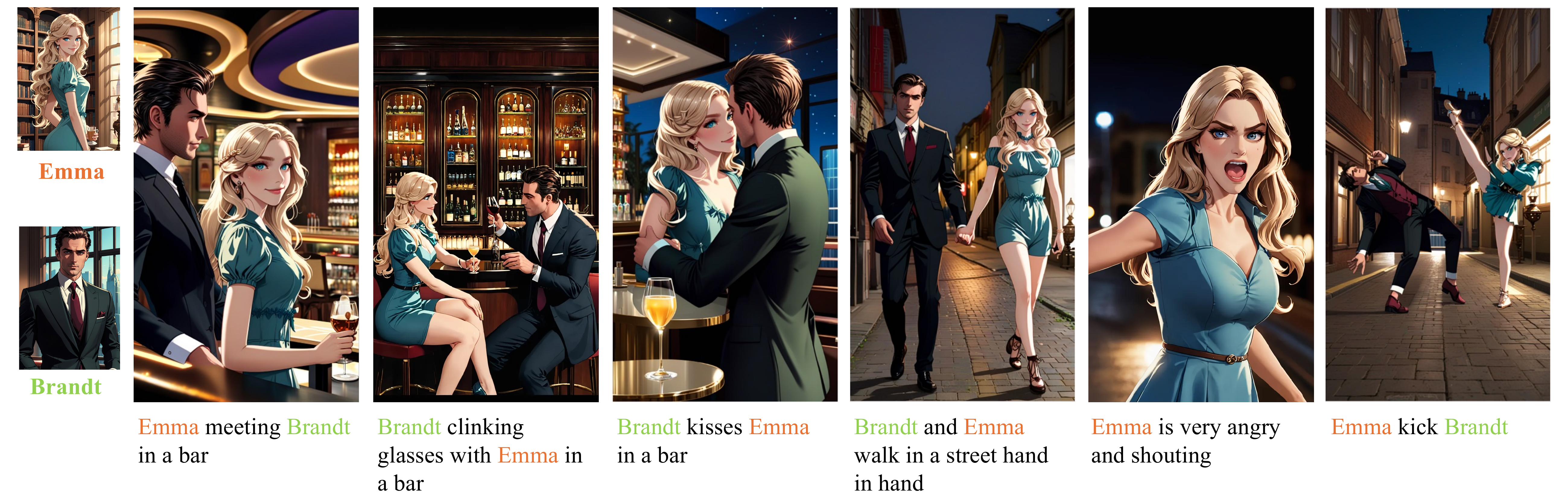}
\captionof{figure}{The story of Brant and Emma. Left: portrait of Emma and Brandt; Right, results of our \model.
  }
\label{fig:teaser}
\end{center}
%
]

\begin{abstract}
We present \model, a novel system for generating comic-style image sequences with consistent characters and expressive gestures. \model addresses two key challenges: (1) maintaining character identity and costume consistency across frames, and (2) producing diverse and vivid character gestures. Our approach integrates a retrieval-based character assignment module, which aligns characters in textual prompts with reference images, and a regional character injection mechanism that embeds character features into specified image regions. Experimental results demonstrate that \model effectively generates engaging comic narratives with coherent characters and dynamic interactions. The source code will be publicly available to support further research in this area.
\end{abstract}

\section{Introduction}







The rapid advancement of text-to-image generation models has opened new avenues for creative applications, notably in the realm of comic creation. These models have demonstrated remarkable capabilities in generating diverse and high-quality images from textual prompts. However, crafting coherent comic narratives presents unique challenges that general-purpose models are ill-equipped to handle. Specifically, they often struggle with maintaining character consistency across panels and generating dynamic, expressive gestures that convey the story’s emotional depth.

\begin{figure*}[t]
    \centering
    \includegraphics[width=0.9\linewidth]{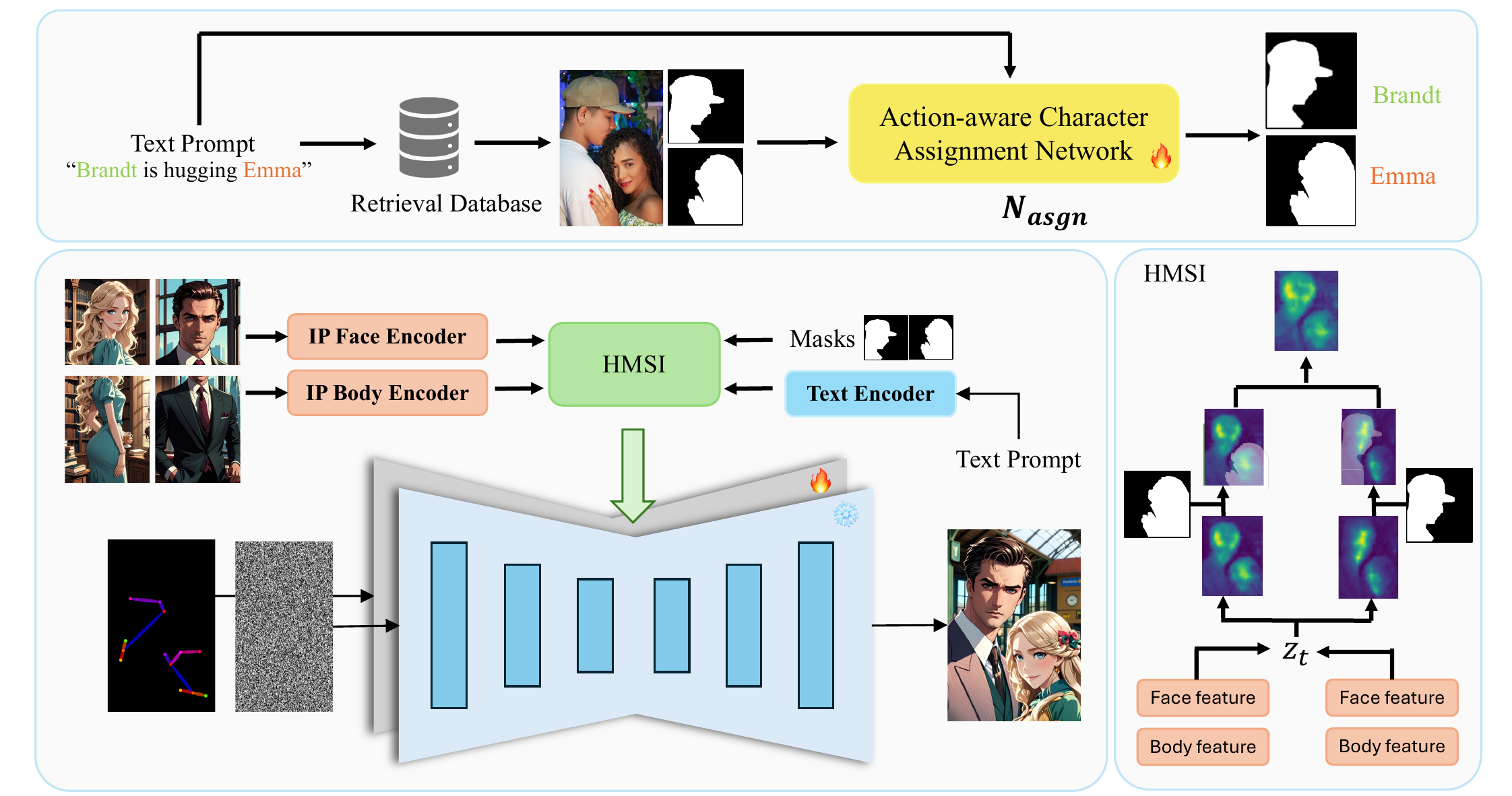}
\captionof{figure}{\textbf{Architecture of the Retrieval Augmented Comic Image Generation.}
  }
    \vspace{-0.5cm}
    \label{fig:arch}
\end{figure*}

In this technical report, we introduce \model, a novel system designed to address two critical challenges in comic image generation:
1) Character Identity and Costume Consistency: Ensuring that characters retain their distinctive features and attire throughout the comic, thereby preserving narrative coherence.
2) Gesture Diversity and Vividness: Generating a wide range of expressive poses and interactions that bring characters to life and enhance storytelling.

To tackle these challenges, \model integrates two key components:
1) Retrieval-Based Character Assignment: This module retrieves reference images from a curated database, aligning characters in the textual prompts with corresponding subjects in the retrieved visuals. This alignment ensures that each character’s identity and costume are consistently represented across different panels.
2) Regional Character Injection: Leveraging the retrieved references, this component injects specific character features into designated regions of the generated images. By focusing on regional attributes, the system maintains character consistency while allowing for varied and expressive gestures.

Our experiments demonstrate that \model effectively generates engaging comic stories featuring consistent characters and rich, dynamic interactions. The system’s ability to preserve character identity and enhance gesture diversity marks a significant step forward in automated comic creation.

The source code for \model will be made publicly available to facilitate further research and development in this area.

\section{Related Works}
\subsection{Retrieval Augmented Generation}
Retrieval-augmented approaches have emerged as a pivotal methodology to enhance the controllability and fidelity of generative models. In the context of image generation, these techniques primarily address hallucination phenomena and the generation of unseen real-world entities by retrieving concrete visual concepts as reference anchors, thereby improving generation plausibility and precision. Current retrieval-augmented image generation methods can be categorized into training-based and training-free paradigms. Notable implementations include Re-Imagen\cite{chen2022re}, which augments intermediate denoising representations through cross-attention with retrieved visual neighbors to generate novel objects absent from the model's training corpus; KNN-Diffusion \cite{sheynin2022knn} that integrates retrieved image features during training while leveraging text-based retrieval for feature generation at inference time to enhance textual faithfulness; and FairRAG\cite{shrestha2024fairrag} which concatenates retrieved image encodings with text embeddings to achieve equitable multi-ethnic human generation. In this work, we propose a retrieval-augmented generation framework specifically designed to enhance action diversity and biomechanical plausibility in character animation.

\subsection{Personalized Image Generation Models}
Personalized image generation is an important method to keep character consistency in story generation. Early approaches such as IP-Adapter\cite{ipadapter} and InstantID\cite{wang2024instantid} achieved personalized image generation through CLIP-based facial feature extraction and cross-attention mechanisms for reference image injection into latent representations. While these methods demonstrate effective single-subject customization, they inherently lack multi-character generation capability. Subsequent works addressing multi-subject scenarios have developed distinct technical frameworks: FastComposer\cite{xiao2024fastcomposer} integrates visual information into text embeddings through a sophisticated pipeline combining textual alignment refinement with auxiliary mask loss fine-tuning. UniPortrait\cite{he2024uniportrait} leverages an IP-Adapter-like architecture for identity feature injection, introducing an id routing mechanism to resolve identity blending challenges in multi-ID customization. Mixture-of-Attention (MoA)\cite{wang2024moa} proposes attention-level feature fusion through router networks, while InstantFamily\cite{kim2024instantfamily} combines OpenPose-guided pose estimation with masked identity feature fusion for family portrait generation. However, current multi-subject personalization systems primarily produce group-style portraits with limited action diversity and gestural expressiveness, fundamentally constraining their capacity for narrative visual storytelling.

\section{\label{sec:dataset}Dataset \dataset}
We construct the dataset \dataset for the training and the retrieval of the \model. To enhance the diversity of human actions, we propose a hierarchical keyword construction strategy with multi-level action variations. Specifically, our framework operates through two complementary dimensions: 1) The first layer systematically varies the number of subjects in scenarios. 2) The second layer generates contextually appropriate verbs across different daily scenarios through semantic expansion. This multi-stage approach ultimately produces 1,800 verb-subject pairs that comprehensively cover common daily activities. Based on these keywords, we successfully collected over 110k rich human action images on the Internet.

For dataset preprocessing, the Sapiens\cite{khirodkar2024sapiens} was employed to perform head and body segmentation for each individual within the images, while RTMO\cite{lu2024rtmo} was subsequently applied for precise human pose estimation.

We generate image captions using GPT-4o \cite{openai2024gpt4ocard} with a systematic approach to mitigate potential biases. Specifically, we employ the symbolic notation "Character $x$" (Character 1, Character 2, ...) as a generic person identifier while removing all appearance-related descriptors (including but not limited to clothing attributes) by carefully designed prompt. This abstraction mechanism is implemented to prevent the introduction of gender-related or physical characteristic biases during cross-modal retrieval tasks, and improving robustness in both training and retrieval stage.

\section{Methodology}

We introduce the framework of \model in~\Cref{fig:arch}. 

\subsection{\label{sec:preliminary}Preliminary}

\subsubsection{Stable Diffusion}
Stable Diffusion~\cite{podell2023sdxl} utilizes a pre-trained variational autoencoder with an encoder $\mathcal{E}$ and a decoder $\mathcal{D}$. For an image $x$, the encoder produces an initial latent $z = \mathcal{E}(x)$, and during the diffusion process, noise is incrementally added, creating a noisy latent $z_t$ at each timestep $t \in T$. The transformation for each step is mathematically described as follows:

\begin{equation}
\begin{gathered}
z_{t} = \sqrt{\bar\alpha_{t}} z_{0} + \sqrt{1 - \bar\alpha_{t}} \epsilon, \quad \epsilon \sim \mathcal{N}(0, I)
\end{gathered}
\label{eq:diffusion}
\end{equation}


\noindent where $\bar\alpha_{t}$ is a fixed noise scaling factor, and $\epsilon$ is noise sampled from a standard normal distribution.

Then Stable Diffusion~\cite{podell2023sdxl} employs a U-Net $\epsilon_\theta$ to predict the noise in the latent $z_t$, guided by the text instructions $c_T$. The diffusion model loss $\mathcal(L)_\text{dm}$ of Stable Diffusion~\cite{podell2023sdxl} is as follows: 

\begin{equation}
\mathcal{L}_\text{dm} = \mathbb{E}_{\mathcal{E}(x), c_T, \epsilon, t }\Big[ \Vert \epsilon - \epsilon_\theta(z_{t}, t, c_T)) \Vert_{2}^{2}\Big]
\label{eq:ldm}
\end{equation}

\subsubsection{IP-Adapter}
IP-Adapter \cite{ipadapter} is a personalized image generation method based on decoupled cross attention mechanism, which incorporates separate cross-attention layers for image features. The core components are image encoder $\mathcal{E}_{img}$ and adapter modules $\boldsymbol{W}^\prime$ in cross attention layers. Given the latent query features $\mathbf{Z}$, the text features $\boldsymbol{c}_t$, and the encoded reference image feature $\boldsymbol{c}_i$, output of cross-attention $\mathbf{Z}^{\prime}$ can be defined by the following equation:

\begin{align}
    &\mathbf{Z}^{\prime}=\operatorname{Attention}(\mathbf{Q}, \mathbf{K}, \mathbf{V}) +  \operatorname{Attention}(\mathbf{Q}, \mathbf{K^\prime}, \mathbf{V^\prime})  \\
    &\operatorname{Attention}(\mathbf{Q}, \mathbf{K}, \mathbf{V}) = \operatorname{Softmax}\left(\frac{\mathbf{Q} \mathbf{K}^{\top}}{\sqrt{d}}\right) \mathbf{V}
    \label{eq:ipa_cross} 
\end{align}

where $\mathbf{Q}=\mathbf{Z} \mathbf{W}_q, \mathbf{K}=\boldsymbol{c}_t \mathbf{W}_k, \mathbf{V}=\boldsymbol{c}_t \mathbf{W}_v, \mathbf{K^\prime}=\boldsymbol{c}_t \mathbf{W^\prime}_k, \mathbf{V^\prime}=\boldsymbol{c}_t \mathbf{W^\prime}_v$ are the query, key, and values matrices of the attention operation respectively, and $\mathbf{W}_q, \mathbf{W}_k, \mathbf{W}_v$ are the weight matrices of the textual trainable linear projection layers.

\subsubsection{ControlNet}
ControlNet \cite{zhang2023adding} extends diffusion models by incorporating spatially explicit conditioning inputs (e.g., edge maps, skeleton maps) through parameter-efficient architectural adaptation. Given a pre-trained U-Net block with parameters $\boldsymbol{\Theta}$, it creates two parallel branches:

A \textit{locked copy} preserving original parameters $\boldsymbol{\Theta}$ to maintain generation quality and a \textit{trainable copy} $\boldsymbol{\Theta}_c$ processing conditioning signal $c$ through zero-initialized convolution layers

The conditioned feature $f_c$ is computed as:
\begin{equation}
f_c = \mathcal{C}(c, \boldsymbol{\Theta}_c) + \text{ZeroConv}(\mathcal{C}(c,\boldsymbol{\Theta}_c))
\label{eq:controlnet}
\end{equation}
where $\mathcal{C}$ denotes feature extraction operations and $\text{ZeroConv}(\cdot)$ represents convolution layers initialized with zero weights. This zero-initialization preserves original model behavior at training start.

The final output combines both pathways:
\begin{equation}
f_{\text{out}} = f_{\text{original}}(\boldsymbol{\Theta}) + \lambda \cdot f_c
\end{equation}
where $\lambda$ is a scaling factor controlling conditioning strength.

This architecture enables precise spatial control while maintaining the pretrained model's generative capabilities through gradual conditioning integration via the trainable branch.

\subsection{Retrieval-Based Image Drafting}

\subsubsection{Text-Image Retrieval}
Given a text prompt $\tprompt$, we first perform \textit{Retrieval-Based Image Drafting}, which begins by extracting the CLIP feature

\begin{align}
    \cfeature = \mathrm{CLIP}(\tprompt)
    \label{equation:clip_feature}
\end{align}
where the $\mathrm{CLIP}$ indicates the CLIP~\cite{clip} network.

This feature is then used to retrieve $M$ nearest neighbor $\closest$ from the constructed dataset \dataset $\db$ using simple ranking strategy. By direct vector multiplication, we select $N$ images $I_M$ with top-$M$ highest scores. Then we random choose $\closest$ from $I_N$ to imporve diversity:

\begin{align}
    I_M = mult\_and\_rank(\cfeature|\db) \\
    \closest = random\_choose(\cfeature|\db)
    \label{equation:clip_feature}
\end{align}

As detailed in~\Cref{sec:dataset}, each retrieved instance $\closest$ is associated with structured annotations, including a human skeleton $\closestSkeleton$, a segmentation map $\closestSegMap$, and a set of reference images $\closestRef$ for each individual.

\subsubsection{\label{sec:aaca}Action-aware Character Assignment}
We then introduce the \textit{Action-aware Character Assignment} network ($\aaca$), which aligns each character mentioned in the text prompt with the corresponding spatial region in the retrieved image, based on the semantic understanding of the prompt, as illustrated in~\Cref{equation:aaca}.

\begin{align}
    \order = \aaca(\closest, \tprompt, \closestSegMap)
    \label{equation:aaca}
\end{align}

Specifically, $\aaca$ takes as input the retrieved image $\closest$, the text prompt $\tprompt$, and the segmentation maps $\closestSegMap$ corresponding to individual characters, and predicts a permutation $\order$ that assigns each character in the prompt to the appropriate segmented region in the image. In implementation, we use a multi-modality model Grounding-DINO \cite{liu2024grounding} as $\aaca$ backbone, we simply replace the bounding-box regression head with a classifier head, and finetune it with \dataset, as we have already labeled the character order in the caption process mentioned above.  

\subsection{\label{sec:rci}Regional Character Injection}

In this stage, we introduce the Regional Character Injection module $\rci$, which features the Regional Injection of entangled Identity and Gesture Feature.

\subsubsection{Disentangled Character Feature Extraction}

As depicted in~\Cref{equation:ipa}, given an image of a target character, we employee two independent identity feature extractor $\ipa$ proposed in IP-Adapters~\cite{ipadapter} to extract the identity feature depicted in the image. Specifically, we first crop the face , which is encoded as $\ipface$ by facial-sensitive IP-Adapter, and the remaining body is encoded as $\ipbody$ by general IP-Adapter. Then the disentangled feature are separately injected into the face and body region to maintain the identity-consistency in the generated image.

\begin{align}
    \ipface = N_{\scriptsize \mbox{IP\_face}}(\iRefface) \\
    \ipbody = N_{\scriptsize \mbox{IP\_body}}(\iRefbody)
    \label{equation:ipa}
\end{align}

\subsubsection{Harmonious Multi-Subject Injection}

Now that the Retrieval-Based Image Drafting determines the region and posture of each character, a straightforward idea is to apply the region prompt to inject the character feature $\ipfeature$ into specific region. However, this is not compactible with the global text prompt given by the users. Moreover, the regional prompt lacks global context, leading to unnatural human interaction and thus unrealistic image.

To tackle this problem, we propose the Harmonious Multi-Subject Injection (\msi), which makes use of the original global text prompt, while forcing the generated characters conditioned on the reference images.

\msi is based on a discovery that, although trained with single person data, the IP-Adapter~\cite{ipadapter} is capable of generating multi-subject images, where all the subjects in the image look consistent to the reference image.

Therefore, we propose to generate the multi-subject injection by performing the IP-Adapter several times and then merge the results according to the spatial region predicted by~\Cref{sec:aaca}. 

However, before the final image is generated, it's hard to identify the precise region. To address this, we propose using controlnet \cite{zhang2023adding} to guide the character features to expected region. Specifically, given a reference image $\iRef^i$ as input, where $i\in[1,N]$ and $N$ denotes the number of subjects in the image, we first compute the $\ipfeature^i$ for each subject according to~\Cref{equation:ipa} and \Cref{eq:ipa_cross}, then inject them into the latent code $z_t$, together with the skeleton feature, as illustrated in~\Cref{eq:controlnet}.

\begin{align}
    \Tilde{z}_t^i = z_t^i + \ipfeature^i +\closestSkeleton + \tprompt
    \label{eq:tildez}
\end{align}

\noindent where $\closestSkeleton$ denotes the controlnet feature of full skeleton map from the nearest piece of data. $\ipfeature$ denotes the fused $\ipface$ and $\ipbody$ according to face and body mask provided by retrieval image, which can be formulated as:

\begin{align}
   &\ipfeature^i = M_{face}^i \cdot \ipface^\prime + M_{body}^i \cdot \ipbody^\prime \\ 
   & \ipface^\prime = \operatorname{Attention}(z_t^i, \ipface, \ipface) \\
   &\ipbody^\prime = \operatorname{Attention}(z_t^i, \ipbody, \ipbody) \\
   &\closestSegMap^i = M_{face}^i \cup M_{body}^i
    \label{eq:fuse_face_body_ip} 
\end{align}

The eventual latent code is formulated as:

\begin{align}
    \Tilde{z}_t = \sum_{i\in[1,N]}{\Tilde{z}_t^i \cdot \closestSegMap^i}
    \label{eq:tildez}
\end{align}

\section{Experiments}
\subsection{Implementation Details}

We train our model based on Stable Diffusion XL\cite{rombach2022high}. In IP preserving branch, we use two IP-Adapter variants, for face preserving, we use IP-Adapter-Plus-Face, as it's pretrained more on face dataset, and for clothes preserving, we use IP-Adapter-Plus,  as it's pretrained on general dataset. We use OpenCLIP ViT-H/14\cite{Ilharco_Open_Clip_2021} as the image encoder. We initialize our controlnet from pretrained model. During training, only controlnet's parameters are trainable. Our model is trained for 7k steps on 8 NVIDIA A800 GPU with a batch size of 2 per GPU at resolution of 1344$\times$768. We use AdamW with a learning rate of 1e-6 all over the training process. During inference, we use DPMSolverPlusSDE\cite{lu2022dpm} sampler with 17 steps and set classifier-free guidance to 2.5. 

In Action-aware character assignment network, considering a maximum of 2 characters, we simply set 2 classes in the final classifier head. During training, we initialize the model from original pretrained weights. And finetune the classifier head on our \dataset use AdamW with 1e-4 learning rate and 1e-4 weight decay. We finetune it on 8 NVIDIA A800 GPU with a batch size of 20 per GPU.

\subsection{Qualitative Results}
\begin{figure*}[t]
    \centering
    \includegraphics[width=0.9\linewidth]{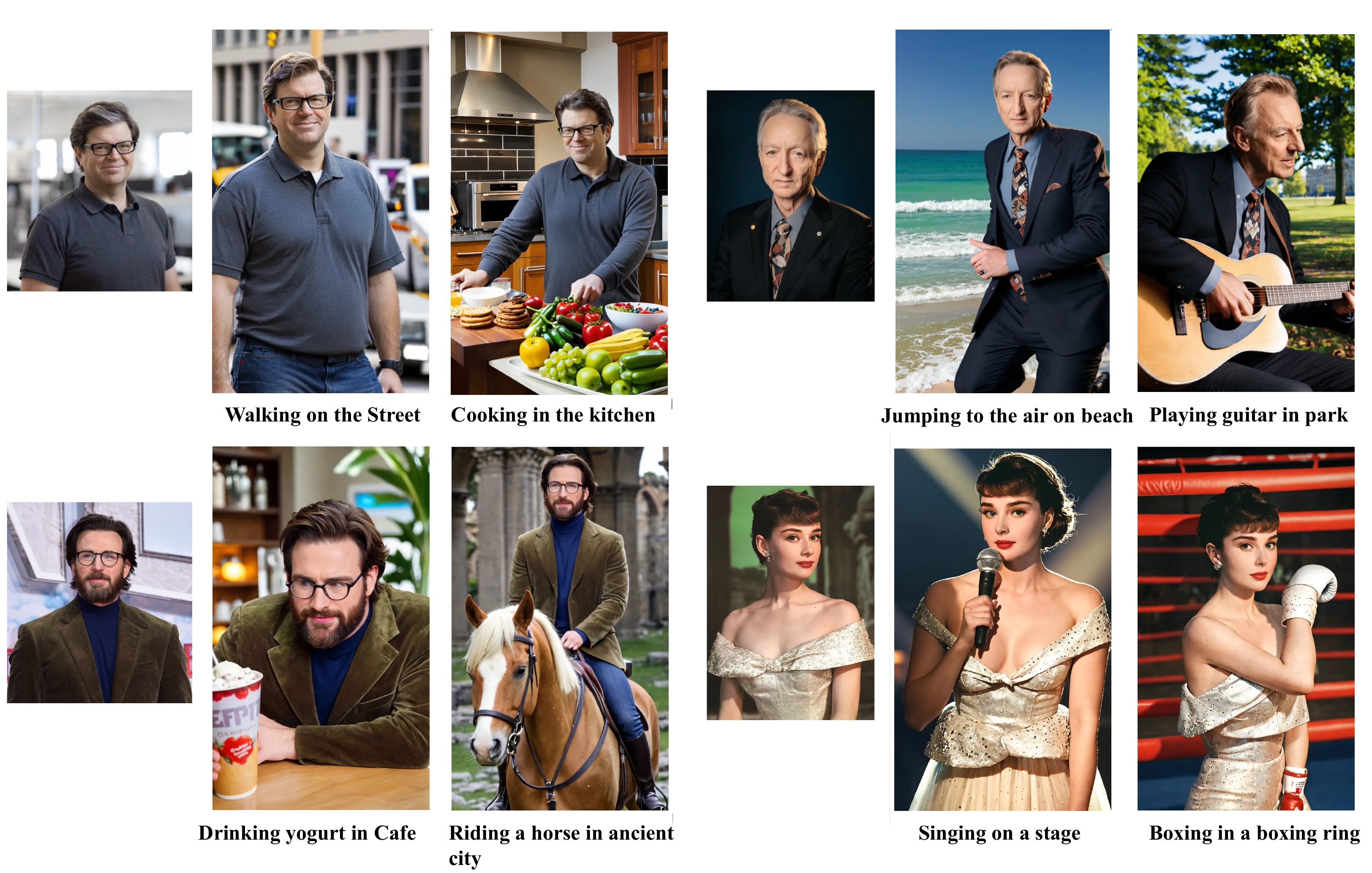}
\captionof{figure}{\textbf{Personalized image generation}.
  }
    \label{fig:single_vis}
\end{figure*}

\paragraph{Personalized image generation}
As shown in \Cref{fig:single_vis}, our model can generate high quality personalized image giving reference image and text control. And as show in last set, because of retrieval-based image drafting, we can generate vivid human action.

\begin{figure*}[t]
    \centering
    \includegraphics[width=0.9\linewidth]{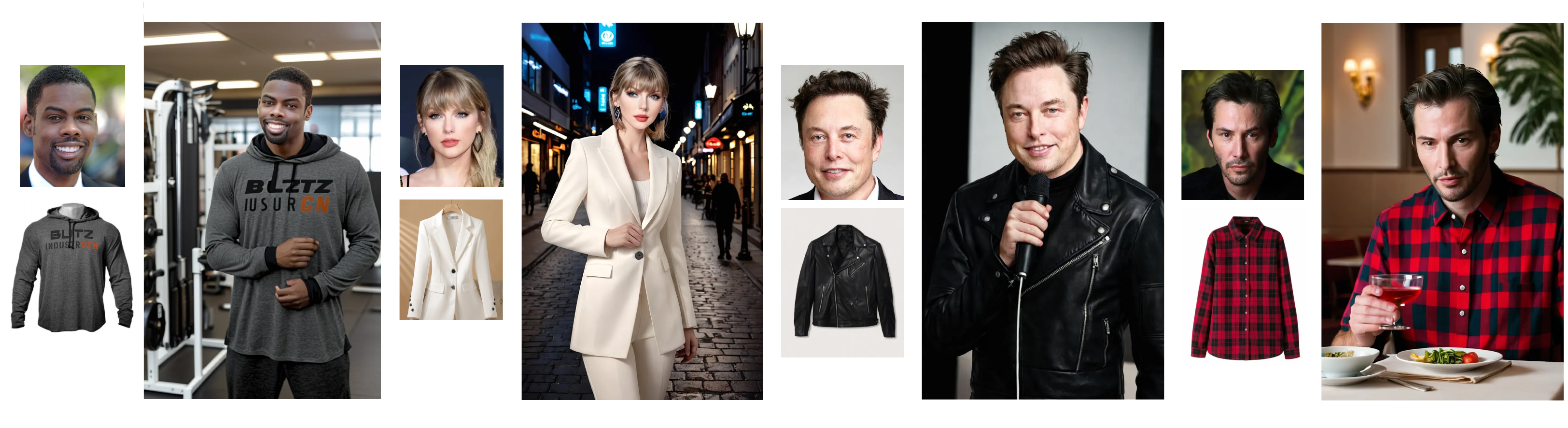}
\captionof{figure}{\textbf{Application of virtual try-on}.
  }
    \vspace{-0.5cm}
    \label{fig:vton}
\end{figure*}

 Due to our utilization of two IP-Adapters in generation, we achieve decoupling facial reference and cloth reference, resulting in application of virtual try-on. Shown in \Cref{fig:vton}

\begin{figure*}[t]
    \centering
    \includegraphics[width=0.9\linewidth]{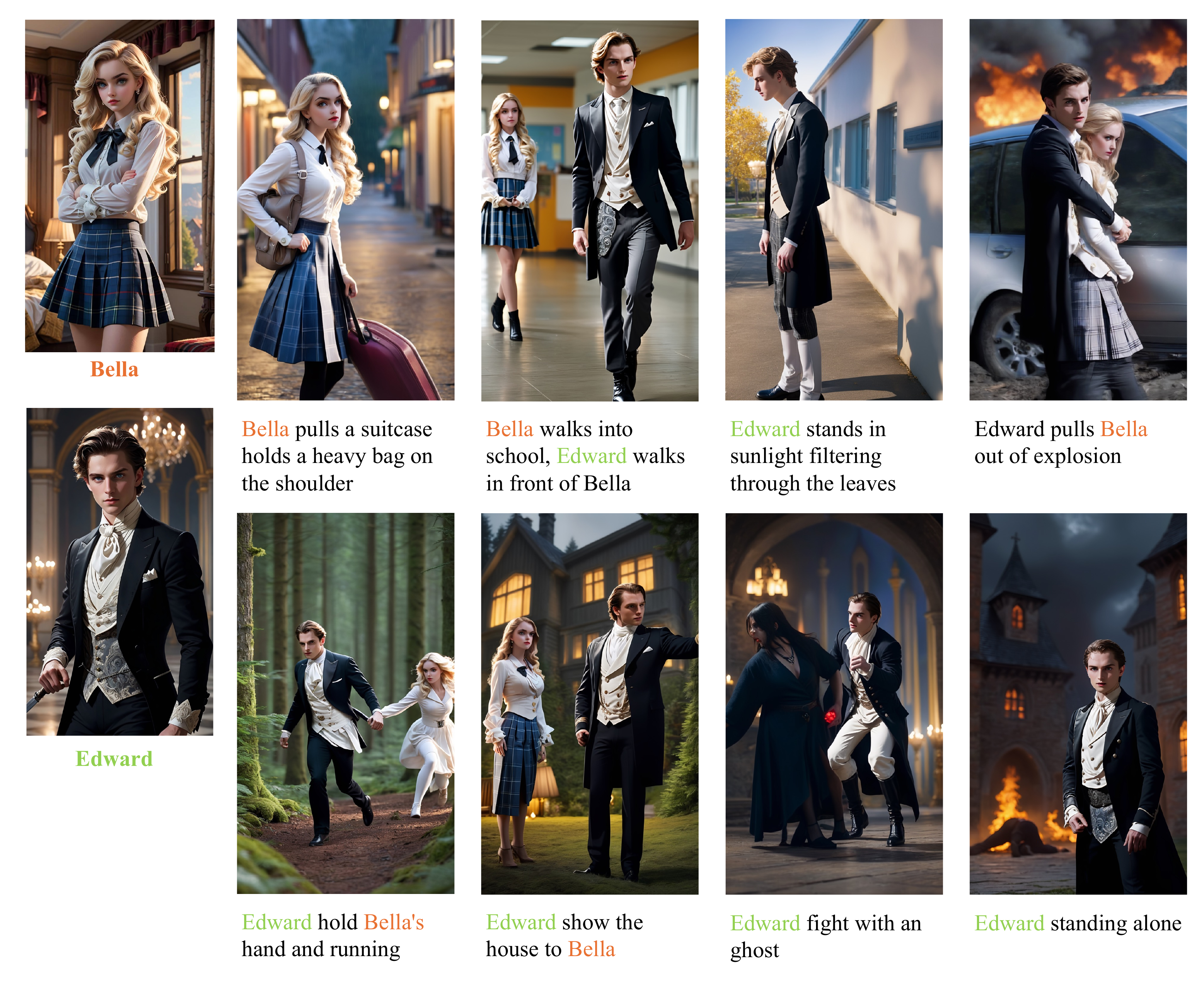}
\captionof{figure}{\textbf{Story 1.} A story between bella and edward, realistic anime style.
  }
    \vspace{-0.5cm}
    \label{fig:story1}
\end{figure*}

\begin{figure*}[t]
    \centering
    \includegraphics[width=0.9\linewidth]{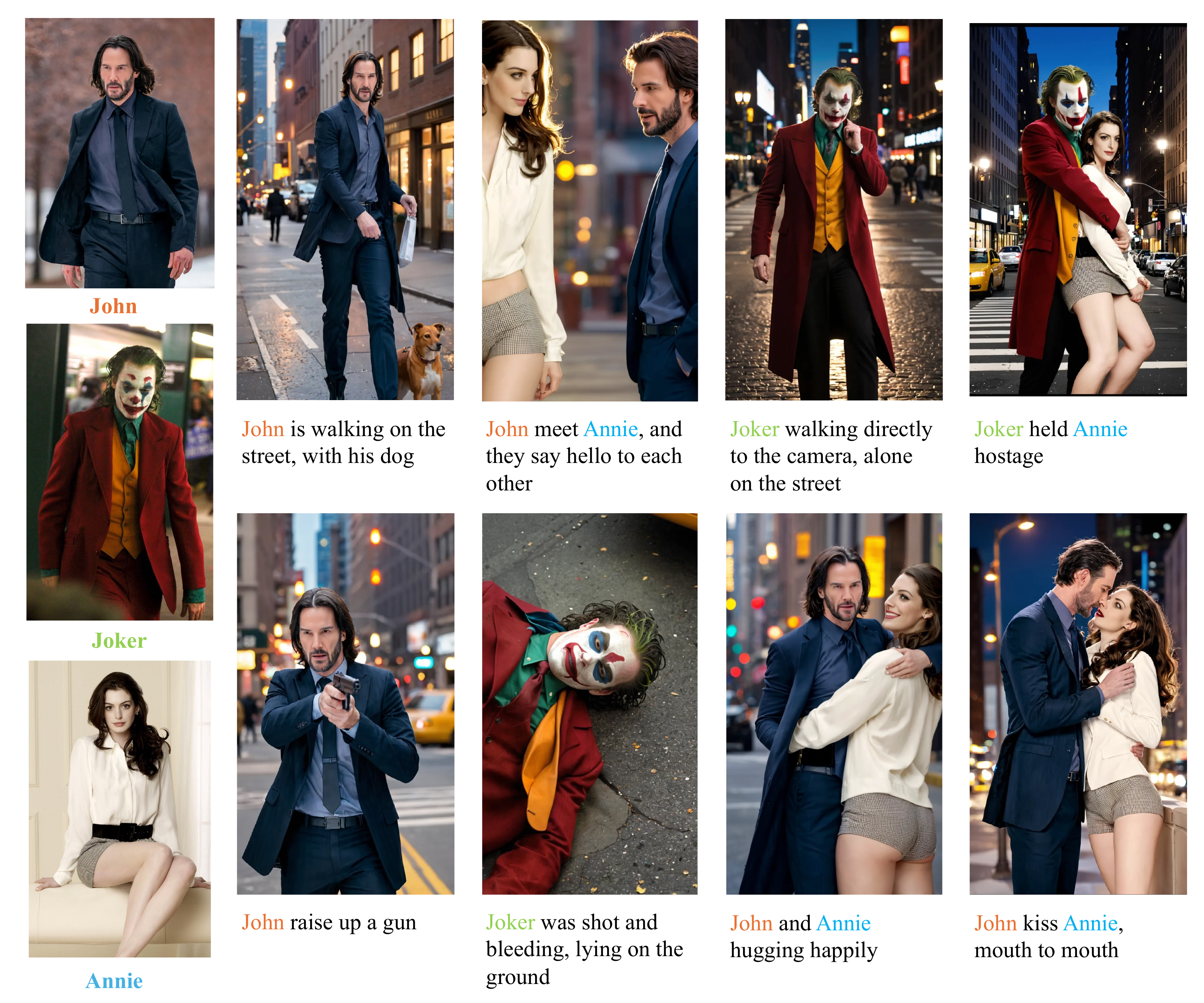}
\captionof{figure}{\textbf{Story 2.} A story between John, Joker and Annie, realistic style.
  }
    \vspace{-0.5cm}
    \label{fig:story2}
\end{figure*}

\begin{figure*}[t]
    \centering
    \includegraphics[width=0.9\linewidth]{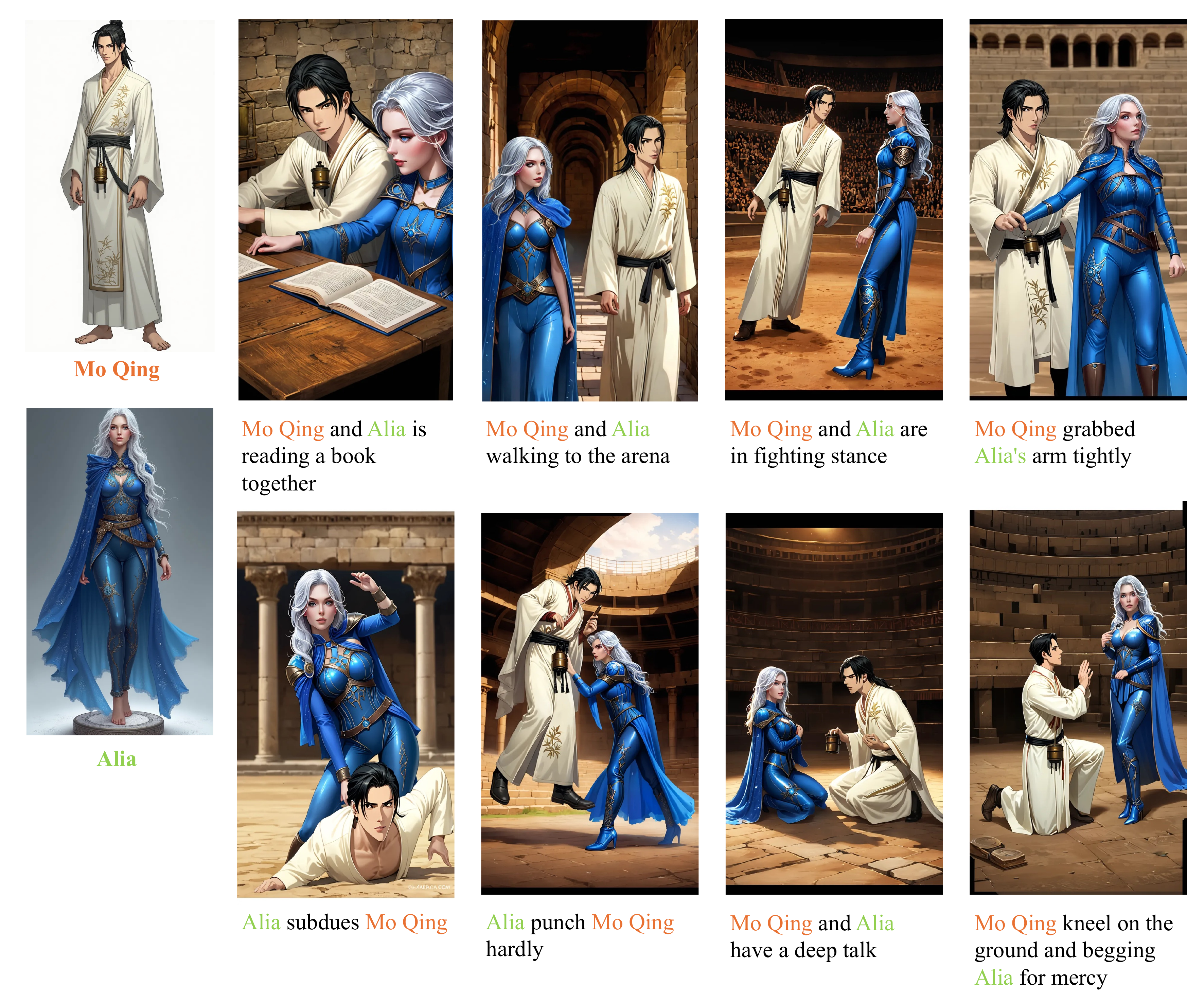}
\captionof{figure}{\textbf{Story 3.} A story between Mo Qing and Alia, comic style.
  }
    \vspace{-0.5cm}
    \label{fig:story3}
\end{figure*}

\paragraph{Story Generation} As shown in \Cref{fig:story1}, \ref{fig:story2}, \ref{fig:story3}, given a text prompt, characters' appearance, our model can generate vivid human interaction  image while preserving facial feature, clothing. In story 1, we tell a story similar to the movie "Twilight" between Bella and Edward. We can see that the character consistency is well maintained across the whole story, while complex and vivid posture are achieved by retrieval mechanism. And in story 3, thanks to retrieval, we can generate many relative complex human pose image for original backbone.

{\small
\bibliographystyle{plainnat}
\bibliography{ref}
}

\end{document}